\documentclass[conference]{IEEEtran}
\IEEEoverridecommandlockouts
\usepackage{xcolor,soul,framed} %,caption
\colorlet{shadecolor}{yellow}
\usepackage{tabularray}
\usepackage{diagbox}
\usepackage{cite}
\usepackage{amsmath,amssymb,amsfonts}
\usepackage{algorithmic}
\usepackage{graphicx}
\usepackage{textcomp}
\usepackage{xcolor}
\usepackage{array}
\usepackage{longtable}
\usepackage{booktabs}
\def\BibTeX{{\rm B\kern-.05em{\sc i\kern-.025em b}\kern-.08em
    T\kern-.1667em\lower.7ex\hbox{E}\kern-.125emX}}
\begin{document}

\title{
Study on the Impacts of Hazardous Behaviors on Autonomous Vehicle Collision Rates Based on Humanoid Scenario Generation in CARLA
% Study on impacts of collision rates based on realistic and random humanoid behaviors in CARLA
% Humanoid Behaviors in CARLA: Providing Realistic Hazard and Randomness to Driving Environment
% \\
\thanks{*Corresponding authors: Min Hua and Quan Zhou}
}

\author{\IEEEauthorblockN{1\textsuperscript{st} Longfei Mo}
\IEEEauthorblockA{\textit{Department of Engineering} \\
\textit{University of Birmingham}\\
Birmingham, UK\\
lxm083@alumni.bham.ac.uk}
\and
\IEEEauthorblockN{2\textsuperscript{nd} Min Hua*}
\IEEEauthorblockA{\textit{Department of Engineering} \\
\textit{University of Birmingham}\\
Birmingham, UK\\
mxh623@student.bham.ac.uk}
\and
\IEEEauthorblockN{3\textsuperscript{rd} Hongyu Sun}
\IEEEauthorblockA{\textit{Department of Engineering} \\
\textit{University of Birmingham}\\
Birmingham, UK\\
h.m.xu@bham.ac.uk}
\and
\IEEEauthorblockN{4\textsuperscript{th} Hongming Xu}
\IEEEauthorblockA{\textit{Department of Engineering} \\
\textit{University of Birmingham}\\
Birmingham, UK\\
 hxs406@student.bham.ac.uk}
 \and
\IEEEauthorblockN{5\textsuperscript{th} Bin Shuai}
\IEEEauthorblockA{\textit{State Key Laboratory of Automotive Safety and Energy} \\
\textit{Tsinghua University}\\
Beijing, China\\
shuaib@tsinghua.edu.cn}
 \and
\IEEEauthorblockN{6\textsuperscript{th} Quan Zhou*}
\IEEEauthorblockA{\textit{Department of  Engineering} \\
\textit{University of Birmingham}\\
Birmingham, UK\\
q.zhou@bham.ac.uk}
}

\maketitle

\begin{abstract}
Testing of function safety and Safety Of The Intended Functionality (SOTIF) is important for autonomous vehicles (AVs). It is hard to test the AV's hazard response in the real world because it would involve hazards to passengers and other road users. This paper studied on virtual testing of AV on the CARLA platform and proposed a Humanoid Scenario Generation (HSG) scheme to investigate the impacts of hazardous behaviors on AV collision rates. The HSG scheme breakthrough the current limitation on the rarity and reproducibility of real scenes. By accurately capturing five prominent human driver behaviors that directly contribute to vehicle collisions in the real world, the methodology significantly enhances the realism and diversity of the simulation, as evidenced by collision rate statistics across various traffic scenarios. Thus, the modular framework allows for customization, and its seamless integration within the CARLA platform ensures compatibility with existing tools. Ultimately, the comparison results demonstrate that all vehicles that exhibited hazardous behaviors followed the predefined random speed distribution and the effectiveness of the HSG was validated by the distinct characteristics displayed by these behaviors.

% \hl{add a key conclusion about the 'impacts'}

% to provide a valuable tool for testing and validating autonomous driving algorithms. 

\end{abstract}


\section{Introduction}
Autonomous vehicles (AVs) have emerged as a groundbreaking technology that holds the potential to revolutionize transportation and reshape various industries. With advancements in sensor technology, artificial intelligence, and computing power, autonomous driving (AD) can operate and navigate without human intervention, offering numerous benefits such as improved safety, enhanced mobility, and increased efficiency. However, the development and deployment of AVs involve complex challenges and considerations. Ensuring the safety of the intended functionality (SOTIF) remains a significant concern, requiring the development of robust algorithms capable of handling complex scenarios.

The complexity and unpredictability of real-world driving scenarios demand a comprehensive evaluation framework that goes beyond traditional road testing.
Nonetheless, scenario simulation plays a crucial role in complementing physical testing by reducing the reliance on extensive real-world trials, mitigating risks, and accelerating the development cycle of AD technologies \cite{b1, b2}. Currently, there are three mainstream scenario generation methods \cite{b3}: data-driven, adversarial, and knowledge-based. The first one relies on collecting a dataset and utilizing density estimation models to generate scenarios. However, it may struggle to generate hazard scenarios accurately since they are rare occurrences in real-life driving datasets. As for the second one, the generation process considers the ego vehicle and incorporates an adversarial learning framework. Consequently, the algorithm becomes more likely to generate scenarios that fail to adhere to realistic traffic rules and constraints. The knowledge-based method involves using pre-defined rules by experts or integrating external knowledge during scenario generation \cite{b4}. However, it lacks fidelity and transferability, making it less effective.

In addition, existing scenario platforms offer relatively fixed test scenarios based on pre-defined scenarios \cite{b5, b6} from organizations like the National Highway Traffic Safety Administration (NHTSA). Examples of such platforms include CARLA Scenario Runner \cite{b7}, DI-Drive Casezoo \cite{b8}, and SMARTS \cite{b9}. The advantages of this approach are its ease of implementation and intuitive nature. However, there are some drawbacks to consider, including a lack of variety and randomness in the scenarios and challenges in selecting appropriate scenario parameters.

To address the limitations of existing methods for SOTIF, this paper presents the humanoid scenario generation (HSG) that centered around the humanoid agents in the CARLA simulation environment. By generating scenarios with randomized parameters and realistic behavior, this approach enhances the authenticity, diversity, and unpredictability of simulated scenes. The proposed approach effectively expedites the creation of diverse scenes from a single safety-critical scenario, providing authenticity, randomization, parameterization, modularity, and ease of integration within the CARLA platform, providing researchers and developers with a valuable tool for testing and validating autonomous driving algorithms.

The remainder of the paper is organized as follows. Section II designs the humanoid hazardous behaviors in Carla Simulator. The validation is conducted in Section III, followed by results and discussion in Section IV, and conclusions are drawn in Section V.

\begin{figure}[ht]
\centering
\includegraphics[width=1\linewidth]{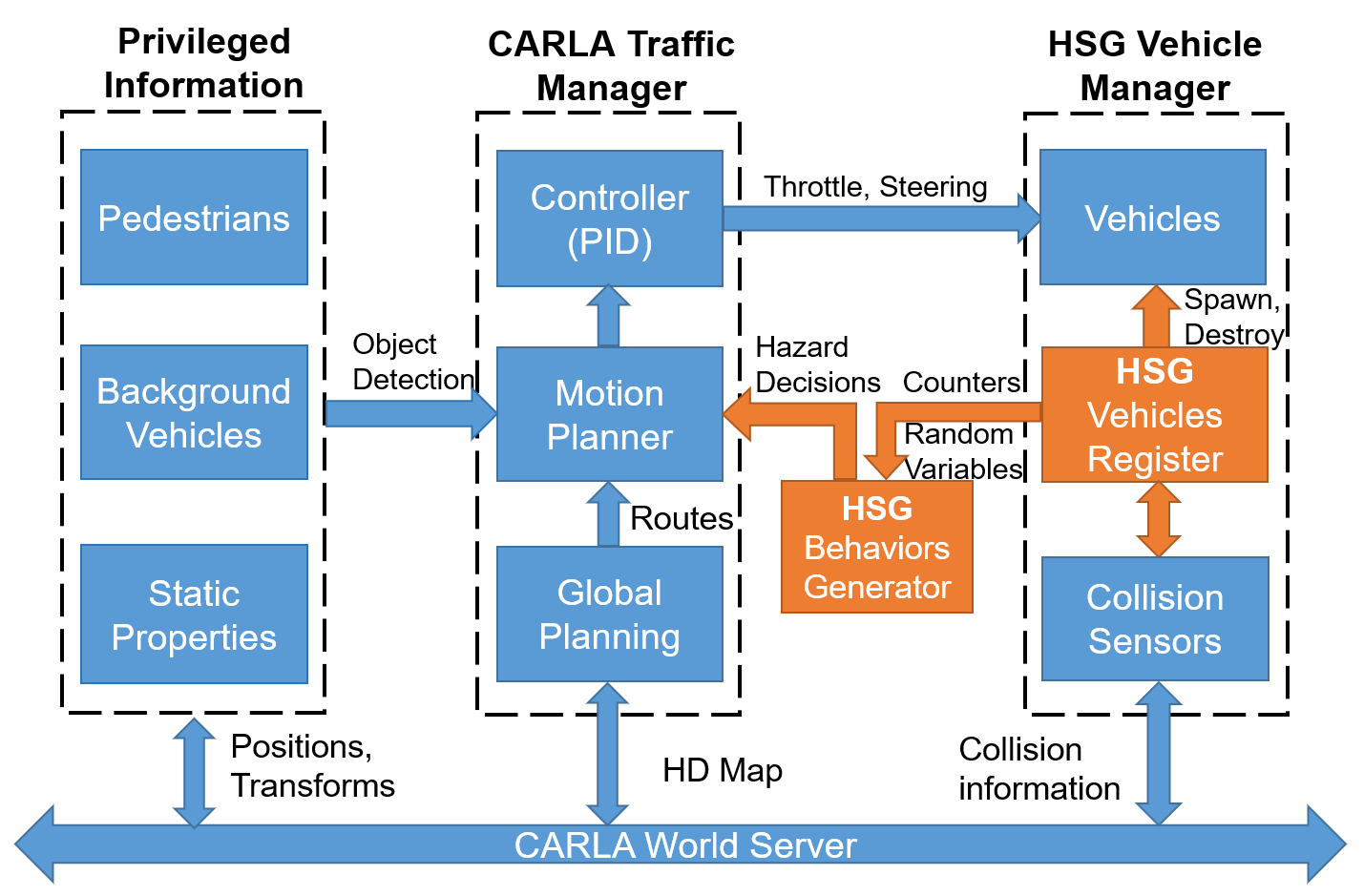}
\caption{The overall framework: the orange modules represent the
HSG, while the blue modules are the built-in modules in CARLA\cite{b10}.}
\label{fig0}
\end{figure}

\section{Humanoid Scenario Generation}
In this study, we will consider five of the most hazardous behaviors observed in real-life scenarios: speeding, impeding, crimping and occupying, drunk and drug driving, and distracted driving\cite{b11}. Instead, classifying driving behavior by basic driving motions\cite{b12, b13} and driving styles\cite{b14, b15} has been much studied today\cite{b16}. 

Research concerning the hazardous driving classification methodology, as expounded in this study, remains relatively scarce. However, it is crucial to note that these dangerous driving behaviors directly contribute to most traffic accidents in real-world scenarios\cite{b17}. Consequently, employing this classification approach in developing models for dangerous driving behaviors holds significant potential in generating critical safety-oriented scenarios. By comparison, the current and above-mentioned methodologies are more general and less likely to yield hazardous and realistic situations.

To accurately simulate these dangerous behaviors, we have modified the decision-making and control mechanisms of the vehicles using the API provided by the traffic manager. The reason for utilizing the traffic manager API stems from its inclusion as a high-performance, built-in module of CARLA developed in C++. This powerful module facilitates the generation of many vehicles with motion planning and control functionalities, thereby enabling the creation of highly realistic traffic scenarios, as shown in Figure \ref{fig0}. Leveraging the traffic manager's capabilities, we can concurrently simulate and control hundreds of vehicles, creating a dynamic and challenging traffic environment.

\subsection{Speeding behaviors}
Speeding vehicles operating within the module will exceed designated speed limits and maintain a shorter bumper-to-bumper distance than regular vehicles, resulting in increased risks. This behavior introduces several hazards, as the higher speed requires a longer braking distance, making it more difficult to stop the vehicle in case of an accident. Thus, the probability of accidents occurring is significantly heightened. In the module, we will control the dangerous behavior of speeding vehicles by a variable parameter called overspeeding ratio $R_{speed}$.

\subsection{Impeding behaviors}
In contrast to speeding vehicles, an impeding vehicle within the module operates at a significantly lower speed than the designated speed limit and maintains a larger bumper-to-bumper distance than a regular vehicle. This behavior has the potential to cause traffic congestion and disruptions on the road, as other vehicles are compelled to wait for the impeding vehicle to pass. Furthermore, the presence of an impeding vehicle may prompt other vehicles to attempt overtaking or bypassing maneuvers, thereby further elevating the risk of accidents. To regulate the hazard posed by impeding vehicles, we will employ a variable parameter known as the impeding ratio $R_{imped}$.

\subsection{Occupying and crimping}
Vehicles that drive outside their designated lanes or occupy multiple lanes present a substantial risk of colliding with other vehicles, significantly increasing the likelihood of accidents. In the module, the ego vehicle is programmed to drive at a restricted speed to ensure safer and more controlled operations. Additionally, the distance $d_{\mathrm{offset}}$ between the vehicle and the centerline will be periodically adjusted based on a predetermined time interval $t_{\mathrm{next}}$ stored in the counter. This approach leads to aggressive driving where the vehicle is always dynamically adjusted within a certain offset range.

\begin{equation}
\begin{aligned}
\label{equ1}
d_{\mathrm{offset}} \sim 0.5 &N(-0.85, 0.5^2) + 0.5 N(0.85, 0.5^2),\\ 
&d_{\mathrm{offset}} \in [-1.7, 1.7]
\end{aligned}
\end{equation}

\begin{equation}
\label{equ2}
t_{\mathrm{next}} \sim U(2, 10)
\end{equation}

% \begin{figure}[ht]
% \centering
% \includegraphics[width=0.8\linewidth]{Figures/figure4.png}
% \caption{The bimodal distribution of $d_{offset}$.}
% \label{fig1}
% \end{figure}

Moreover, the $d_{\mathrm{offset}}$ will follow a bimodal distribution, 
% in Figure \ref{fig1}, 
which allows $d_{\mathrm{offset}}$ to get values with a higher probability of keeping vehicles on both sides of the lane rather than in the center. $t_{\mathrm{next}}$ represents the time of transition between vehicle states, which takes values with equal probability in a certain interval and will therefore obey a continuous uniform distribution.

\subsection{Drunk and drug driving}
The driver's diminishing reflexes and impaired judgment can lead to a loss of control over the vehicle or an inability to accurately assess road conditions and the positions of other vehicles\cite{b18}, resulting in traffic accidents. In the module implementation, drivers exhibit a probability denoted as $P_d$ to disregard traffic signals and signs (e.g., stop signs) and engage in speeding or impeding behaviors. These behaviors are manifested at different times with a speed ratio denoted as $r_d$. Consequently, the speed ratio $r_d$ will be adjusted periodically based on a predefined time interval denoted as $t_d$, which is stored in the counter.

\begin{equation}
\label{equ3}
r_d \sim N(1, 0.25^2),\quad r_d \in [0.5, 1.5]
\end{equation}

\begin{equation}
\label{equ4}
t_{d} \sim U(4, 10)
\end{equation}

The value of $P_d$ follows a Gaussian distribution with a mean of 0. This indicates that the vehicle has the highest probability of driving at a speed close to the speed limit, with a certain possibility of driving at significantly higher or lower speeds. Similarly, the time intervals for state changes, represented by $t_d$, follow a continuous uniform distribution within a specified range.

\subsection{Distracted driving}
Distractions, such as cell phone usage, eating, or engaging in conversation with passengers\cite{b19, b20, b21}, can divert the driver's attention. During these distracted moments, the driver's perception and decision-making abilities may lapse, potentially leading to situations where the vehicle veers off the road or fails to respond promptly to obstacles. Reports indicate that the typical duration of such distractions ranges from 1.5 to 3 seconds.

\begin{equation}
\label{equ5}
t_{\mathrm{loss}} \sim N(1.5, 0.5^2),\quad t_{\mathrm{loss}} \in [1, 3]
\end{equation}

\begin{equation}
\label{equ6}
t_{\mathrm{start}} \sim U(t_{\mathrm{loss}}, 30-t_{\mathrm{loss}})
\end{equation}

In the module implementation, the ego vehicle ceases to generate new control parameters once every $t_{\mathrm{cycle}}$ seconds when the distraction begins, starting at time $t_{\mathrm{start}}$. The previously generated control parameters are then repeatedly applied to the vehicle for a duration of $t_{\mathrm{loss}}$ seconds. The value of $t_{\mathrm{cycle}}$ is a pre-determined random variable selected before the simulation(e.g. 30s), while the values of $t_{\mathrm{start}}$ and $t_{\mathrm{loss}}$ are variables that change over time during the simulation. Specifically, the value of $t_{\mathrm{start}}$ follows a continuous uniform distribution, and the value of $t_{\mathrm{loss}}$ follows a Gaussian distribution.

\section{Simulation Design}
% \subsection{Environment settings}
The majority of simulation variables, such as vehicle model selection, generation location, traffic light timing, and random distribution values, are determined randomly using Python's random library and NumPy's random module. By using the same seed for each case, consistent random variables are employed, resulting in reproducible experimental outcomes. The default seed for the experiments is set to 10, but certain subsequent experiments may necessitate the use of multiple distinct seeds.

\subsection{Hazardous behaviors validation}
\begin{figure}[ht]
\centering
\includegraphics[width=0.5\textwidth]{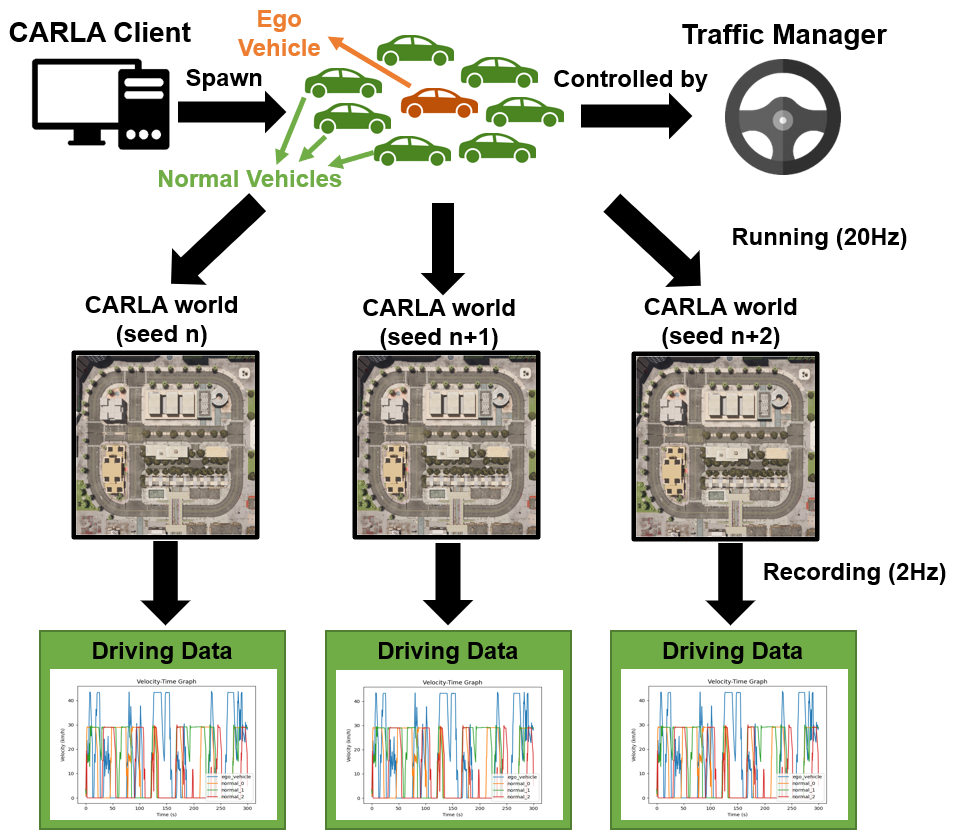}
\caption{In the simulation, we will use the CARLA client to generate 50 autonomous vehicles controlled by the traffic manager randomly and set one of them as the ego vehicle and apply hazard behaviors to it. Each hazard behavior agent will be tested in one or a few CARLA’s worlds with different seeds and a sampling frequency of 2Hz per simulation.}
\label{fig2}
\end{figure}

In this case, we will verify hazardous behaviors generated by ego vehicles, including speeding, impeding, crimping and occupying, drunk and drug driving, and distracted driving. The driving data of the ego vehicle will be recorded and used to analyze whether it exhibits the corresponding behaviors, as depicted in Figure \ref{fig2}. The speeding vehicle will have a $R_{\mathrm{speed}}=150$\%, resulting in a maximum speed of 45km/h, while normal vehicles have a maximum speed of 30km/h. The velocity of the ego vehicle and three other normal vehicles will be drawn. For the impeding vehicle, with $R_{\mathrm{imped}}=50$\%, its maximum speed will be 15km/h. The velocity of the ego vehicle, three normal vehicles, and their average following distance will be recorded. The parameters of other hazardous behaviors will follow the design in Section II, as the recorded data setting are presented in Table \ref{tab1}.

\begin{table}[]
\caption{Overall recorded driving data during the simulations}
\begin{center}
\begin{tabular}{|c|c|c|}
\hline
\textbf{Behaviors} & \textbf{Graphs}      & \textbf{Tables}                                                              \\ \hline
Speeding           & velocity-time        & \begin{tabular}[c]{@{}c@{}}Mean Bumper-to-Bumper \\ Distance\end{tabular} \\ \hline
Impeding           & velocity-time        & \begin{tabular}[c]{@{}c@{}}Mean Bumper-to-Bumper \\ Distance\end{tabular} \\ \hline
Crimp occupy       & offset distance-time & -                                                                            \\ \hline
Drunk Drug         & velocity-time        & Red light running ratio                                                      \\ \hline
Distracted         & Throttle-time        & -                                                                            \\ \hline
\end{tabular}
\end{center}
\label{tab1}
\end{table}

\begin{figure*}[ht]
\centering
\includegraphics[width=0.8\linewidth]{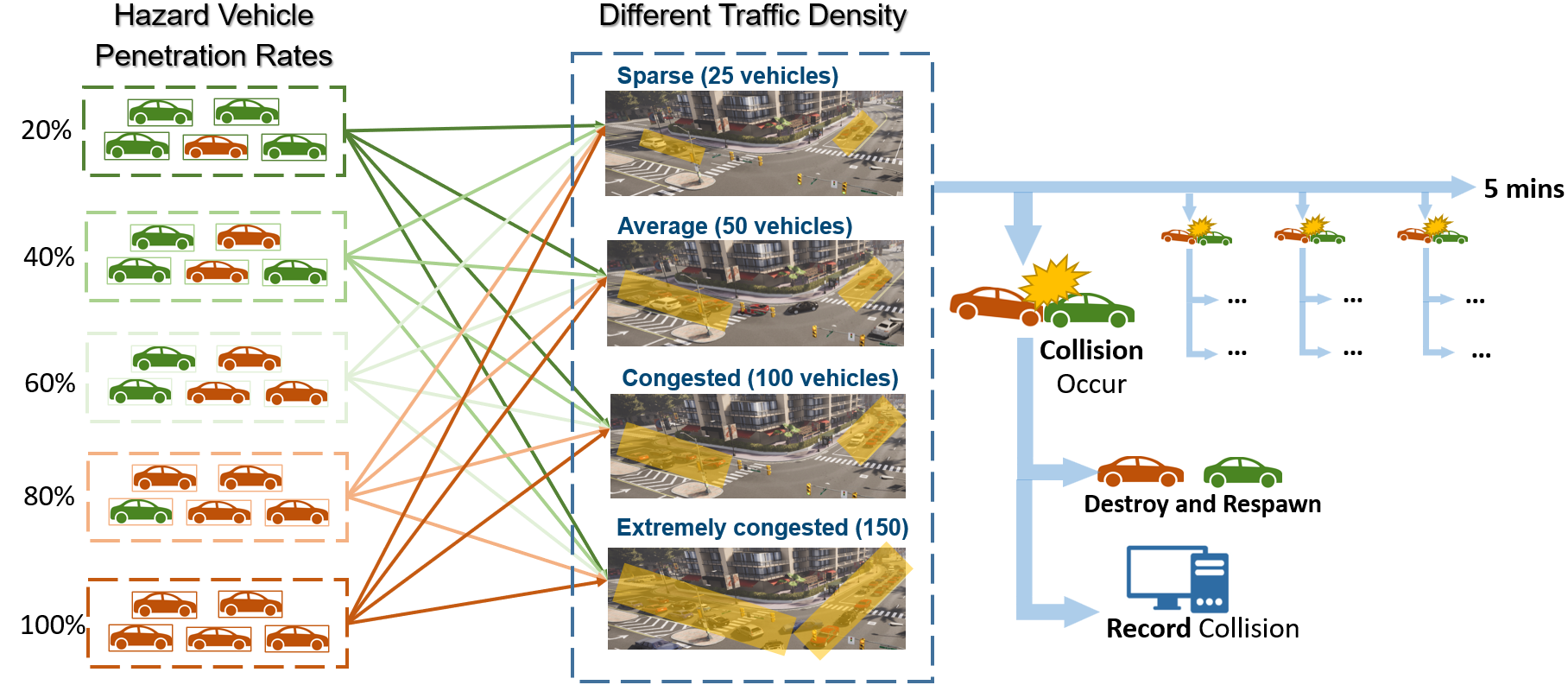}
\caption{The simulation will run 20 times for 5 different hazard vehicle penetration rates (from 20\% to 100\%) and 4 different traffic densities (yellow blocks are used to show the different vehicle densities at the same intersection). Each time the simulation will run for 5 minutes and record all the collision information, and the ratio of five behaviors in total hazardous vehicles is uniform distribution)}
\label{fig3}
\end{figure*}

\subsection{Hazardous behaviors in traffic environment validation}
In this case, the impact of hazardous vehicles on traffic in urban scenarios was investigated by testing the collision rates of traffic simulation in CARLA map Town10. Four traffic densities will be considered in the simulation: 25, 50, 100, and 150, representing sparse, average, congested, and especially congested, respectively. Furthermore, replacing 20\%, 40\%, 60\%, 80\%, and 100\% of all vehicles into hazardous vehicles one at a time. Given the traffic density and penetration rate, there is a total of 20 sets of simulations, each with a seed of 10, and a test time of 5 minutes. The calculation of the penetration rate in each scenario is shown below.

\begin{equation}
\label{equ7}
\textit{Penetration rate} = \frac{N_{hazard}}{N_{total}} 
\end{equation}
Where $N_{hazard}$ is the number of hazardous vehicles and $N_{total}$ is the total number of vehicles in test driving environments.

Collisions are recorded during the simulation, and crashed vehicles are promptly replaced to prevent road obstructions and ensure accurate simulation results. The selected penetration range of 20\%-100\% is justified for specific reasons:

1) At 0\% penetration, where no dangerous vehicles are present, the traffic can run for 2 hours without any collision incidents. Therefore, testing between 0\%-20\% penetration is deemed unnecessary.

2) It is crucial to assess the system's capability to handle the highest level of danger. Hence, a maximum penetration rate of 100\% is required.

3) Although a penetration rate exceeding 40\% for dangerous vehicles in traffic is already highly hazardous in real-life scenarios, the short duration of our single test (5 minutes) may not generate a sufficient number of collisions within that timeframe to adequately evaluate the level of danger.

Considering these factors, the chosen penetration range of 20\%-100\% allows for comprehensive testing of the system's performance and safety capabilities across various scenarios.

\section{Results and Discussion}
\subsection{Discussion with the five hazardous behaviors}

\subsubsection{Speeding vehicles}

\begin{figure*}[ht]
\centering
\includegraphics[width=0.9\linewidth]{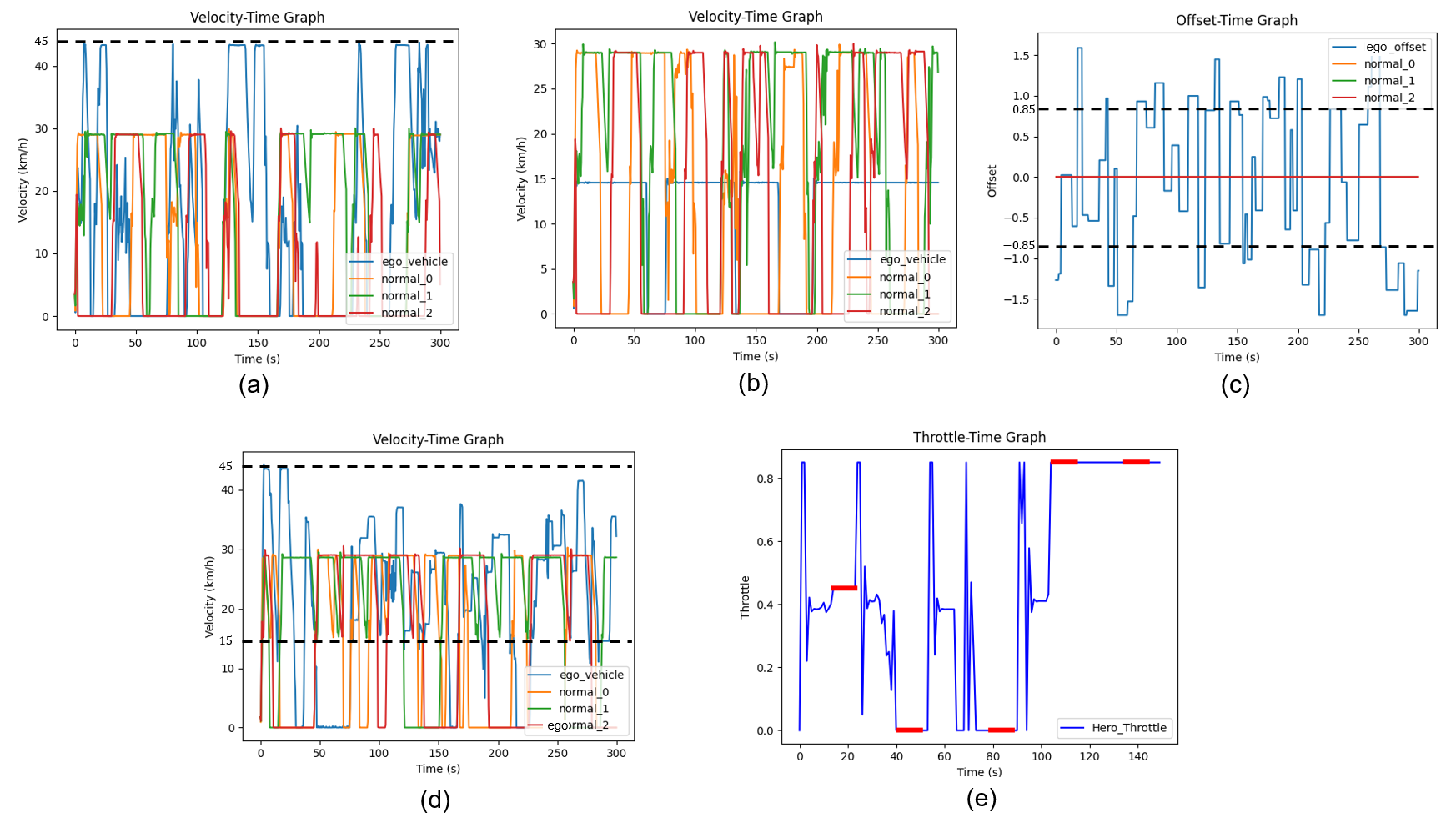}
\caption{(a) shows the v-t graph of speeding vehicles. (b) shows the v-t graph of impeding vehicles. (c) shows the $d_{\mathrm{offset}}$-t graph of crimp and occupy vehicles. (d) shows the v-t graph of drunk and drug vehicles. (e) shows the throttle-time graph of distracted vehicles.}
\label{fig4}
\end{figure*}

In Figure \ref{fig4}(a), the ego vehicle consistently maintained a maximum speed of 45 km/h compared to 30 km/h for other normal vehicles, indicating persistent speeding behavior of the ego vehicle throughout the simulation. The ego vehicle's smaller mean bumper-to-bumper distance compared to other normal vehicles in the speeding group in Table \ref{tab2} suggests a shorter reaction time and braking distance. Speeding behavior increases the danger level for the ego vehicle and poses higher risks to itself and other vehicles on the road.

\subsubsection{Impeding vehicles}
In Figure \ref{fig4}(b), the ego vehicle consistently maintained a maximum speed of 15 km/h, indicating persistent impeding behavior. The ego vehicle's larger mean bumper-to-bumper distance compared to other normal vehicles in the impeding group in Table \ref{tab2} suggests it not only traveled at a slower speed but also maintained a greater distance from leading vehicles. As a result, the ego vehicle is more likely to impede or block traffic under this behavior.

\begin{table}
\centering
\caption{Mean bumper-to-bumper distance for Speeding and Impeding vehicles}
\label{tab2}
\begin{tabular}{|c|c|c|c|l|} 
\hline
\diagbox{Vehicles}{Tests~} & \begin{tabular}[c]{@{}c@{}}Test 1\\ (seed 10)\end{tabular} & \begin{tabular}[c]{@{}c@{}}Test 2\\ (seed 11)\end{tabular} & \begin{tabular}[c]{@{}c@{}}Test 3\\ (seed 12)\end{tabular} & \begin{tabular}[c]{@{}l@{}}Mean \\ Distance\end{tabular}  \\ 
\hline
\multicolumn{5}{|c|}{Speeding Groups}                                                                                                                                                                                                                                         \\ 
\hline
Ego vehicle                & \textbf{14.04 m}                                            & \textbf{13.98m}                                            & 14.87 m                                                     & \textbf{14.30 m}                                           \\ 
\hline
Normal vehicle 1           & 18.17 m                                                     & 20.40 m                                                     & \textbf{13.65 m}                                            & 17.41 m                                                    \\ 
\hline
Normal vehicle 2           & 18.22 m                                                     & 15.88 m                                                     & 18.61 m                                                     & 17.57 m                                                    \\ 
\hline
Normal vehicle 3           & 17.51 m                                                     & 14.57 m                                                     & 16.94 m                                                     & 16.34 m                                                    \\ 
\hline
\multicolumn{5}{|c|}{Impeding Groups}                                                                                                                                                                                                                                         \\ 
\hline
Ego vehicle                & \textbf{15.76 m}                                            & \textbf{11.70 m}                                            & \textbf{11.76 m}                                            & \textbf{13.07 m}                                           \\ 
\hline
Normal vehicle 1           & 12.87 m                                                     & 9.43 m                                                      & 9.09 m                                                      & 10.46 m                                                    \\ 
\hline
Normal vehicle 2           & 13.41 m                                                     & 10.22 m                                                     & 10.04 m                                                     & 11.22 m                                                    \\ 
\hline
Normal vehicle 3           & 15.08 m                                                     & 10.52 m                                                     & 10.68 m                                                     & \multicolumn{1}{c|}{12.09 m}                               \\
\hline
\end{tabular}
\end{table}

\subsubsection{Occupying and crimping vehicles}
In Figure \ref{fig4}(c), The $d_{\mathrm{offset}}$ values of the main vehicle occur more frequently around the two mean values $\mu_1 $ and $\mu_2$, which indicates that the occupying behavior of the ego vehicle basically fits the bimodal distribution set in the hazardous behavior generator.

\subsubsection{Drunk and drug vehicles}
In Figure \ref{fig4}(d), the maximum speed of the ego vehicle is distributed with random values between the maximum speeding and impeding speed, and the average probability of the ego vehicle running a red light in the test also converges to 50\% as the number of tests increases in Table \ref{tab3}. Therefore, we can generally determine that in the experiment, the behavior of the main car does conform to the setting of the hazardous behavior generator.

\subsubsection{Distracting driving}
In Figure \ref{fig4}(e), the red line representing the loss of control appears every 30s and when it occurs, the control of the ego vehicle maintains the same control value for $t_{\mathrm{loss}}=10s$. This means that the ego vehicle fully complies with the settings in the hazardous behavior generator to produce distracted driving behavior.

\begin{table}
\centering
\caption{The red-light running ratio of drunk and drug vehicles}
\label{tab3}
\begin{tblr}{
  row{even} = {c},
  row{3} = {c},
  row{5} = {c},
  cell{1}{1} = {r=2}{},
  cell{1}{2} = {c=5}{c},
  cell{1}{7} = {r=2}{},
  vlines,
  hline{1,3-6} = {-}{},
  hline{2} = {2-6}{},
}
                                      & \textbf{Seeds} &             &             &             &             & \textbf{Mean}  \\
                                      & \textbf{10}    & \textbf{11} & \textbf{12} & \textbf{13} & \textbf{14} &                \\
Running red lights                    & 5              & 4           & 3           & 7           & 8           & -              \\
Total lights                          & 8              & 12          & 9           & 11          & 13          & -              \\
{Running red lights\\ percentage(\%)} & 62.5           & 33.3        & 33.3        & 63.6        & 61.5        & \textbf{50.84} 
\end{tblr}
\end{table}

\begin{figure}[ht]
\centering
\includegraphics[width=0.8\linewidth]{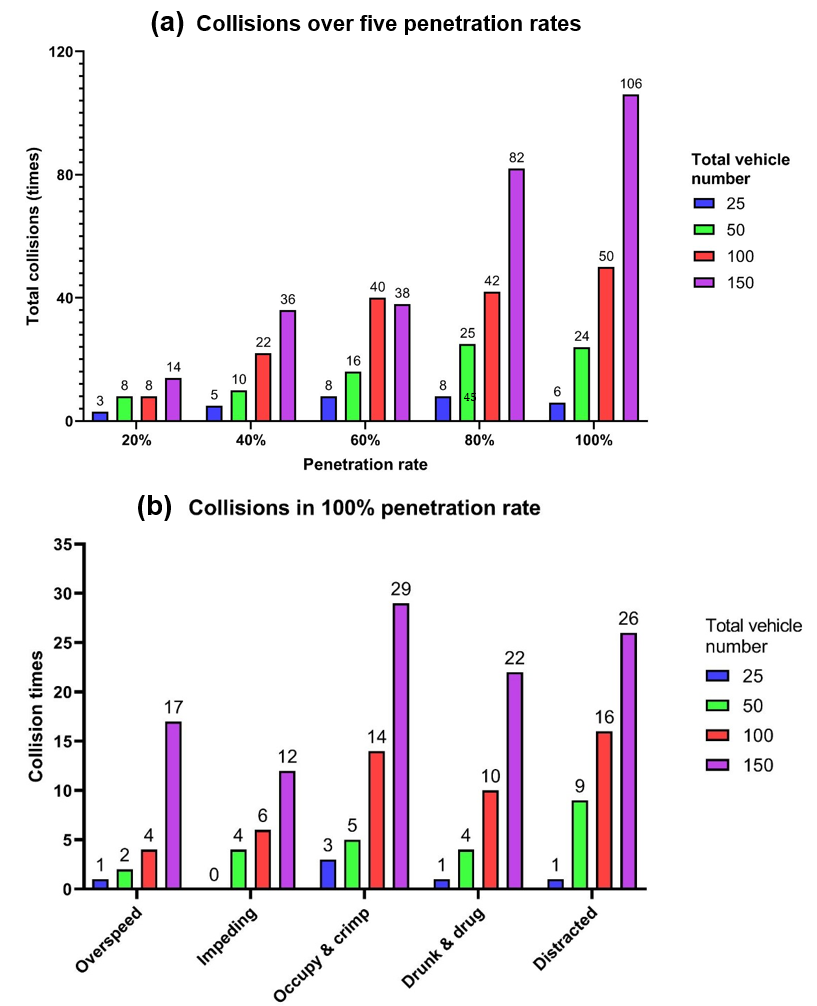}
\caption{(a) the total number of collisions in the traffic environment with different penetration and vehicle density. (b) the number of collisions generated by different vehicle densities and hazardous behaviors, respectively, at 100\% penetration. }
\label{fig5}
\end{figure}

\subsection{Discussion with the hazardous behaviors in traffic environment }

Figure \ref{fig5}(a) shows that hazardous vehicles indeed have a worse impact on the traffic environment and that crash rates increase with penetration and vehicle density, which also indicates that the danger of the driving environment increases significantly with these two factors. Also, we can find that the crash rates are more similar for penetration rates between 40\% and 60\% and between 80\% and 100\%.

In Figure \ref{fig5}(b), collisions increase with vehicle density from 25 to 150. Among the risky behaviors analyzed, distracted driving, occupying, and crimping exhibit the highest collision rates, followed by drunk and drug behavior. Speeding and impeding have relatively low collision rates.

Distracted driving, occupying and crimping behaviors contribute to more collisions than other risky behaviors in urban environments. These behaviors lead to increased lane incursions and impact surrounding vehicles. Aggressive control during accelerating, cornering, and overtaking impairs nearby vehicles' judgment. In high-density scenarios, reduced vehicle distance limits reaction time, increasing the likelihood of collisions.

\section{Conclusion and future works}

This paper implements a humanoid agent in CARLA with five hazardous behaviors. Multiple simulation scenarios examine the effectiveness of the hazardous behavior generator and quantify the hazards introduced by different hazardous driving agents. Based on the findings obtained from the experimental results, the following conclusions can be drawn:

1) Speeding vehicles exhibited a mean bumper-to-bumper distance 16.37\% shorter than that of normal vehicles while impeding vehicles maintained a 16.12\% higher bumper-to-bumper distance. Consequently, these vehicles face an increased risk of collisions or traffic blockages, respectively.

% 1)	Speeding vehicles exhibited an mean bumper-to-bumper distance 16.37\% shorter than normal vehicles while impeding vehicles maintained a 16.12\% higher bumper-to-bumper distance. Consequently, these agents face an increased risk of collisions or traffic blockages, respectively.
2)	All vehicles displaying hazardous behaviors adhered to the predefined random speed distribution. The validity of the hazardous behavior generator was confirmed by the distinctive characteristics exhibited by these behaviors.

3) Crimping and distracted vehicles had the highest collision frequency, with 29 and 26 collisions, respectively. Drunk and speeding vehicles had intermediate collision numbers while impeding vehicles had the lowest collision rate. This pattern persisted at vehicle densities above 50, emphasizing the increased risk associated with crimping and distracted behaviors in such environments.
% 3)	Crimping and distracted vehicles experienced a higher frequency of collisions, with the most extreme cases resulting in 29 and 26 collisions, respectively. Drunk and speeding vehicles followed with intermediate collision numbers, while impeding vehicles demonstrated the lowest collision rate. This pattern remained consistent at vehicle densities above 50, highlighting the heightened danger posed by crimping and distracted behaviors in such environments.

Future work considers the integration of the humanoid perception module with the hazardous behavior generator module, encompassing coupling at decision-making and motion-planning levels.

\end{document}